\documentclass[pmlr]{jmlr}

\usepackage{longtable}
\DeclareMathOperator*{\argmin}{arg\,min}

 \usepackage{booktabs}
 \usepackage{multirow, svg, todonotes, boldline, bbm, makecell, wrapfig, adjustbox}
 
\usepackage[load-configurations=version-1]{siunitx} 

\makeatletter
\def\set@curr@file#1{\def\@curr@file{#1}} 
\makeatother

\theorembodyfont{\upshape}
\theoremheaderfont{\scshape}
\theorempostheader{:}
\theoremsep{\newline}

\jmlrvolume{182}
\jmlryear{2022}
\jmlrworkshop{Machine Learning for Healthcare}

\title[KeyClass for Automated Text Classification]{Classifying Unstructured Clinical Notes via Automatic Weak Supervision}

\author{\Name{Chufan Gao \nametag{\thanks{Authors contributed equally to this research.}}}
       \Email{chufang@andrew.cmu.edu}\\ \vspace{-3mm}
       \AND
       \Name{Mononito Goswami \nametag{$^ * $} }
       \Email{mgoswami@andrew.cmu.edu}\\ \vspace{-3mm}
       \AND
       \Name{Jieshi Chen}
       \Email{jieshic@andrew.cmu.edu}\\ \vspace{-3mm}
       \AND
       \Name{Artur Dubrawski}
       \Email{awd@andrew.cmu.edu}\\ 
       \addr Auton Lab, School of Computer Science, Carnegie Mellon University, 
       Pittsburgh, PA, USA
} 

\begin{document}

\maketitle

\begin{abstract}
  Healthcare providers usually record detailed notes of the clinical care delivered to each patient for clinical, research, and billing purposes. Due to the unstructured nature of these narratives, providers employ dedicated staff to assign diagnostic codes to patients' diagnoses using the International Classification of Diseases (ICD) coding system. This manual process is not only time-consuming but also costly and error-prone. Prior work demonstrated potential utility of Machine Learning (ML) methodology in automating this process,  but it has relied on large quantities of manually labeled data to train the models. Additionally, diagnostic coding systems evolve with time, which makes traditional supervised learning strategies unable to generalize beyond local applications. In this work, we introduce a general weakly-supervised text classification framework that learns from \textit{class-label descriptions only}, without the need to use any human-labeled documents. It leverages the linguistic domain knowledge stored within pre-trained language models and the data programming framework to assign code labels to individual texts. We demonstrate the efficacy and flexibility of our method by comparing it to state-of-the-art weak text classifiers across four real-world text classification datasets, in addition to assigning ICD codes to medical notes in the publicly available \texttt{MIMIC-III} database.
\end{abstract}

\section{Introduction}

The Electronic Health Record (EHR) system is a digital version of a patient’s paper chart. EHRs are almost-real-time, patient-centered records that contain patient history, diagnoses, procedures, medications, and more in an easily accessible format. Since the Health Information Technology for Economic and Clinical Health (HITECH) Act was signed into law in 2009 \citep{menachemi2011benefits}, adoption rates of these systems have steadily increased. \citet{adler2017electronic}, who analyzed survey data collected by American Hospital Association found that EHR adoption rates were at 80\% in 2017, twice the rate in 2008. With higher adoption rates comes a rising challenge: data processing and analysis of unstructured clinical text. Natural language free-texts are regularly recorded in the form of radiology or discharge notes and are used for diagnostic, research, and billing purposes. 

To be studied and managed adequately, cohorts of patients with similar clinical characteristics need reliable phenotype labels. However, specific phenotype data is seldom available compared to other EHR data, like clinical texts ~\citep{venkataraman2020fastag}. In practice, diagnostic codes are among the most common proxies to true phenotypes. Due to the unstructured nature of clinical notes, providers often employ trained staff and/or third-party vendors to help assign diagnostic codes using coding systems such as the International Classification of Diseases (ICD)~\citep{moriyama2011history}. However, manual assignment of codes is both time consuming and error-prone, with only 60--80\% of the assigned codes reflecting actual patient diagnoses \citep{benesch1997inaccuracy} and significant portion of misjudged severity of conditions and code omissions~\citep{venkataraman2020fastag}. For healthcare providers, billing and coding errors may not only lead to loss of revenue and claim denials, but also federal penalties for erroneous Medicare and Medicaid claims. Thus, there is a clear need for reliable automated classification of unstructured clinical notes.

Prior work introduced the use of Machine Learning (ML) to automatically assign diagnostic codes to clinical notes \citep{venkataraman2020fastag, baumel2018multi, yu2019automatic, xu2019multimodal}. Yet, most of the involved ML models rely on vast quantities of pointillistically labeled training data, which is often unavailable or costly to collect. In addition, coding systems are periodically revised, rendering already labeled data at least partially obsolete. In fact, ICD is currently in its $10$th revision\footnote{ICD-10 version was released in 1992; however, in this paper, we restrict our experiments to assigning ICD-9 codes to better evaluate the performance of our proposed methods vis-\`{a}-vis prior work.}, while its $11$th revision has already been accepted by the World Health Organization and will come into effect on January 2022~\citep{wiki:ICD}. To make matters worse, most providers use their internal coding systems, making traditional supervised ML strategies infeasible to generalize across organizations. 

As a potential remedy, we present \texttt{KeyClass}, a general weakly supervised text classification framework combining Data Programming \citep{ratner2016data} with a novel method of automatically acquiring interpretable weak supervision sources (keywords and phrases) from \textit{class-label descriptions only} without the need to access to any labeled documents. The successful application of \texttt{KeyClass} to solve an important clinical text classification problem demonstrates its potential for making social impact by allowing quick and affordable development and deployment of effective text classifiers. Our primary contributions include:
\begin{itemize}
    \item We introduce a general weakly supervised text classification model \texttt{KeyClass}, and a novel strategy to effectively and efficiently acquire interpretable weak supervision sources for text, to learn highly discriminative text classifiers only from descriptions of classes, without any human-labeled documents.\footnote{The code for \texttt{KeyClass} will be made publicly available at \url{https://github.com/autonlab/KeyClass}.}

    \item We use \texttt{KeyClass} to reliably assign ICD-9 codes to patient discharge notes with no labeled documents and minimal human effort. Experiments on the publicly available \texttt{MIMIC-III} dataset reveal that \texttt{KeyClass} performs comparably to a robust supervised alternative \citep{venkataraman2020fastag} trained using several thousand manually annotated clinical notes.
    
    \item We conduct experiments on $4$ other common multiclass text classification datasets to benchmark \texttt{KeyClass} against previously proposed weakly supervised methods. Results reveal that our model efficiently and effectively creats text classifiers that outperform prior work.
    
    \item To the best of our knowledge, \texttt{KeyClass} is the first to employ data programming for classification in a multiclass multilabel setting. The ICD-9 assignment problem involves assigning \textit{all} relevant codes to each clinical note.
\end{itemize}

\begin{figure}[ht!]
    \centering
    \includegraphics[width=.75\textwidth]{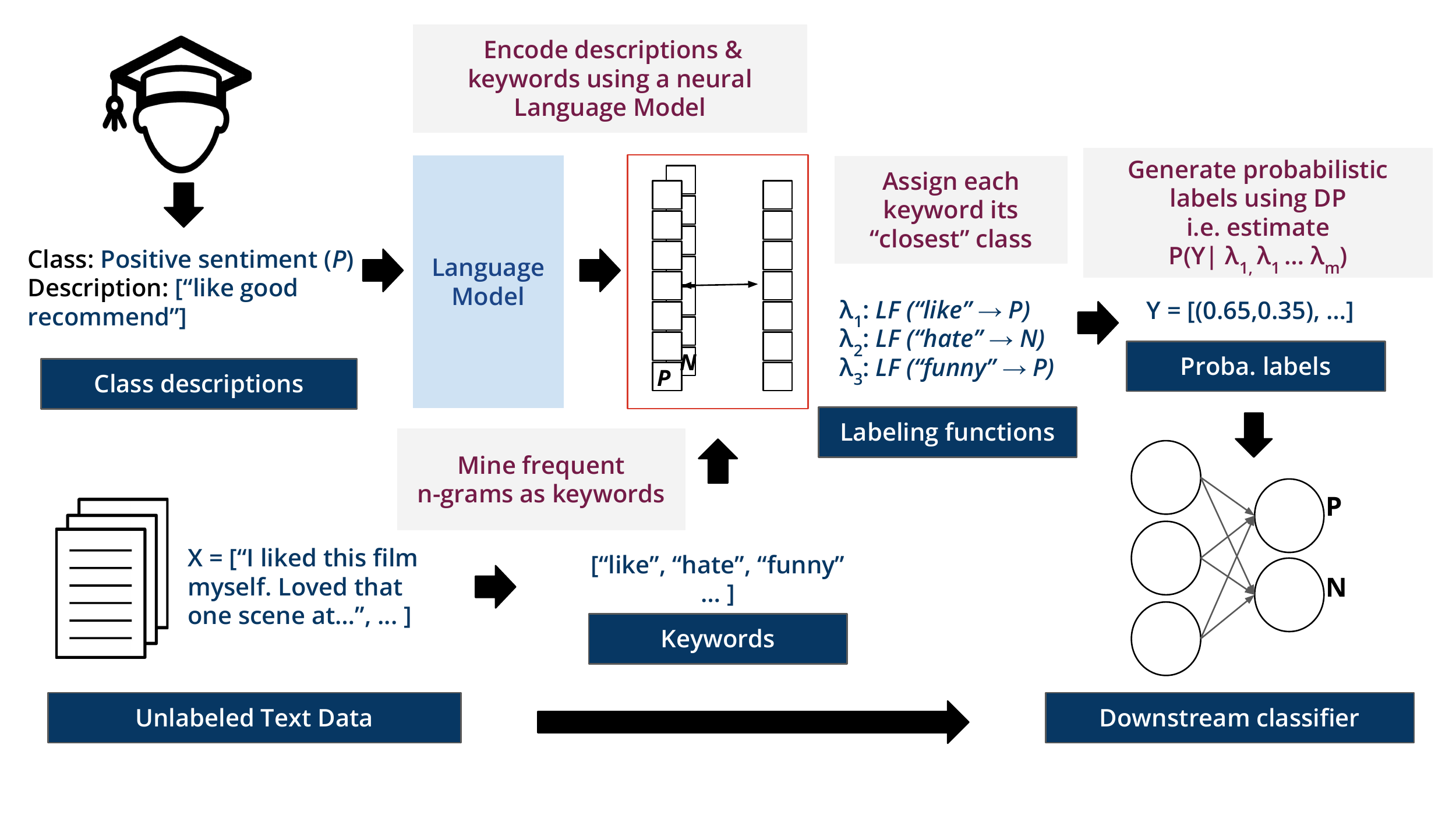}
    \caption{Overview of our methodology. From only class descriptions, \texttt{KeyClass} classifies documents without access to any labeled data. It automatically creates interpretable labeling functions (LFs) by extracting frequent \textit{keywords} and \textit{phrases} that are highly indicative of a particular class from the unlabeled text using a pre-trained language model. It then uses these LFs along with Data Programming (DP) to generate probabilistic labels for training data, which are used to train a downstream classifier \citep{ratner2016data}.}
    \label{fig:outline}
\end{figure}

\subsection*{Generalizable Insights about Machine Learning in the Context of Healthcare}


Managing costs and quality of healthcare is a persistent societal challenge of enormous magnitude and impact on daily lives of all people. Our work targets one very specific aspect of this complex landscape. Our approach proposes a low-cost solution that has the potential to address some of the identified pressing issues with accessibility to affordable yet accurate automated disease coding tools. Our contributions lie in using a novel strategy to efficiently acquire interpretable weak supervision sources from readily available text to learn effective text classifiers without the need for human-labeled data. Results on multiple datasets demonstrate that our method can outperform state-of-the-art baselines in realistic settings, and it can perform comparably to a fully supervised model in an important clinical problem. Our work demonstrates that (1) pre-trained language models can efficiently and effectively inform weakly supervised models for text classification, (2) self-training improves downstream classifier performance, especially when classifiers are initially trained on a subset of the training data, (3) data programming performs on par with simple majority vote when relying on a large number of automatically generated weak supervision sources of similar quality, and (4) key words are excellent sources of weak supervision. 

\section{Prior Work}
\label{sec:priorwork}

\subsection{Assigning ICD codes to Clinical Notes}
ICD code assignment is hard due to the multiclass and multilabel nature of the problem. ICD-9 for instance defines more than 14,000 unique codes for nuanced classification of diseases, symptoms, abnormal findings, etc. Recent revisions of ICD have a much greater number of codes permitting the classification of new and previously known conditions with higher precision and finer granularity~\citep{world1988international}. Moreover, patients may be assigned more than one code depending on their diagnoses. Most studies to date tackle the first problem by either classifying a subset of codes or by grouping them based on the first three characters of four-five character codes. For our study, we follow \citet{venkataraman2020fastag}'s approach and classify general diagnostic categories, of which there are $19$. 

Numerous studies have used ML to tackle the problem of assigning ICD codes to unstructured clinical text. For example, \citet{baumel2018multi} used hierarchical attention bidirectional Gated Recurrent Units (GRU) to tag discharge summaries by identifying sentences related to each label in a hierarchical manner. Outside of the English language domain, \citet{yu2019automatic} investigated the use of a similar hierarchical attention Long Short-Term Memory (LSTM) \citep{lstmHochreiter1997} to assign ICD-10 codes to clinical admission records in Chinese. Some studies have also explored the use multimodal features to improve tagging performance, instead of simply relying on the unstructured text. For example, \citet{xu2019multimodal} proposed an ensemble of modality specific models to predict ICD-10 diagnostic codes. Specifically, they applied a Convolutional Neural Network (Text-CNN) to unstructured free text, an LSTM to semi-structured diagnosis descriptions, and a decision tree to process tabular patient data such as prescriptions, lab and microbiology events. Recently, \citet{venkataraman2020fastag} proposed \texttt{FasTag}, a fully supervised LSTM model which achieved state-of-the-art performance on classifying unstructured patient discharge notes into top-level ICD-9 categories. In addition to ML, some studies have also used information retrieval techniques to support human experts performing tagging. \citet{rizzo2015icd}, for instance, used transfer learning to expand a skewed dataset, while retrieving the top-$K$ relevant codes and passing them to a human expert to improve tagging accuracy.

Another issue in ICD code assignment is that of low-support labels. As can be seen in Table~\ref{tab:mimic_f1}, ICD codes vary significantly in terms of their frequency in data. For example, as many as 70\% of the documents contain \textit{supplementary} ICD codes, whereas only 0.003\% of the documents are assigned \textit{pregnancy or childbirth complication} codes. To combat this issue, \citet{chapman2020automatic} used label descriptions to improve their model's performance on the least represented ICD for Oncology (version 3) (ICD-O-3) codes. They utilized a Bidirectional Encoder Representations from Transformers (BERT)-like \citep{devlin2018bert} encoder and a word-level attention mechanism between input clinical text and textual descriptions of labels, using the output to a model with a customized loss function that favors recall. Through this method, they were able to consistently produce more varied ICD code predictions, while assigning fewer codes to each clinical text and maintaining a high recall and a competitive F1 score. In the rest of the paper, we restrict ourselves to assigning \textit{high-level} ICD-9 codes to discharge notes in the publicly available \texttt{MIMIC-III} database following the same experimental settings as \texttt{FasTag} \citep{venkataraman2020fastag}. Our results (Table~\ref{tab:mimic_f1}) reveal that even state-of-the-art ICD-code classification methods such as \texttt{KeyClass} and \texttt{FasTag} have a hard time predicting low-support categories, which may benefit from further research on classification under high class imbalance.

To the best of our knowledge, all prior work on ICD code assignment utilized fully supervised ML techniques, most of them relying on vast quantities of labeled training data. In this work, we explore the use of our proposed weakly supervised model \texttt{KeyClass} to assign top-level ICD-9 codes to long patient discharge summaries. Its training signal is retrieved automatically from readily available descriptions of the ICD codes, therefore it requires no human-produced supervisory feedback to build effective downstream text classifiers. 



\subsection{Text Classification with Sparse Training Labels}
Weakly supervised text classification aims to classify text documents using cheaper albeit potentially noisier sources of supervision such as keywords. The earliest attempts at weak forms of supervision involved mapping documents and label names to Wikipedia concepts in a semantic space. The semantic relatedness between the labels and the documents are then used to classify text documents \citep{gabrilovich2007computing}. Since these methods do not use any domain specific unlabeled data, relying purely on general knowledge, they are often referred to as \texttt{Dataless} techniques in the literature. Another class of methods use neural models to either generate psuedo documents or detect category indicative words in documents. For instance, the \texttt{WeSTClass} model generates pseudo-documents to pre-train a text classifier followed by self-training on labeled data for model refinement \citep{meng2018weakly}. More recently, \citet{meng2020text} proposed \texttt{LOTClass}, which associates semantically related words to label names and finds the implied category of words via masked category pre diction. Finally, their model self-trains itself on unlabeled documents to improve generalization. 

Inspired by \citet{meng2018weakly, meng2020text}, \texttt{KeyClass} is self-trained on unlabeled training documents using its own highly confident predictions. However, \texttt{KeyClass} differs from prior work in some fundamental ways. First, the foundation of our weak supervision methodology, i.e., frequent keywords and phrases as LFs, is highly interpretable. Secondly, while previously proposed state-of-the-art models are committed to specific language model architectures for linguistic knowledge and representation learning, \texttt{KeyClass} offers a high degree of modularity, enabling end users to adapt the neural language model (encoder) and downstream classifiers to specific problems, such as clinical text classification. Finally, we explore the use of weak supervision for multilabel multiclass classification, a problem which, to the best of our knowledge, has not been tackled by prior work on weak text classification.

\subsection{Weak Supervision for Clinical Text Classification}
Recently, weak supervision has also found use in clinical text classification. For example, \citet{wang2019clinical} developed a manually annotated, rule-based algorithm combined with data programming \citep{ratner2016data} to create weak labels for smoking status, and proximal femur (hip) fracture classification. They used pre-trained word embeddings as deep representation features to train simple ML models for classification. Similarly, \citet{cusick2021using} trained weakly supervised models to detect suicidal ideation from unstructured clinical notes using rule-based labeling functions. 

Thus, prior work on weakly supervised clinical text classification had an explicit dependence on \textit{manually} created rule-based labeling functions. In this work, however, we demonstrate that we can quickly and automatically create simple keyword based labeling functions, with minimal to no human involvement.


\section{Problem Formulation}
Given a collection of $n$ documents $\mathcal{D} = \{d_1\}_{i = 1 \dots n}$, $c$ class labels $\mathcal{C} = \{c_i\}_{i = 1 \dots c}$, and their descriptions $\mathcal{E} = \{e_i\}_{i = 1 \dots c}$, our goal is to first ``probabilistically'' label each document $d_i$ using a label model $\mathcal{L}_\theta$ and then use these labels to train a downstream classifier $\mathcal{M}_\phi$ to assign all relevant class labels $c_i \in \mathcal{S}_i$ to each document $d_i \in \mathcal{D}$, where  $\mathcal{S}_i$ is a set of classes and $\mathcal{S}_i \subseteq \mathcal{C}$. Furthermore, for our experiments, with the exception of ICD-9 code assignment, all problems are single-label multiclass in nature ($\mathcal{S}_i$ is a singleton set). 

 The label model $\mathcal{L}$ parameterized by $\theta$ relies on a set of $m$ \emph{labeling functions} (LF) denoted by $\Lambda = \{\lambda_i\}_{i=1 \dots m}$, where each LF $\lambda_i: \mathcal{D} \rightarrow \mathcal{S} \subseteq \mathcal{C}$, assigns a label $\hat p(c_j \in \mathcal{C} \mid \Lambda)$, to each document $d_i \in \mathcal{D}$. Note that each LF only votes for a single class. In this work, we constrain the set of labeling functions to be simple \textit{keyword-matching rules} of the form:
 \begin{equation}
 \label{eqn:rule}
 \texttt{If } k_i \texttt{ occurs in } d_j \texttt{ then vote } c_k \texttt{ else abstain}
 \end{equation}
where $d_j$ is the $j^{th}$ document, $c_k$ is the $k^{th}$ class, and $k_i$ belongs to a set of keywords or key-phrases $\mathcal{K}$ automatically mined from $\mathcal{D}$ (See Figure~\ref{fig:lf_examples}).

In the following section, we will review data programming methodology used in \texttt{KeyClass} to generate labels from the keyword-matching rules mentioned previously.

\subsection{Data Programming for Weak Text Classification}
\label{sec:dataprogramming}
\begin{figure}[!htb]
    \centering
    \includegraphics[width=0.5\textwidth]{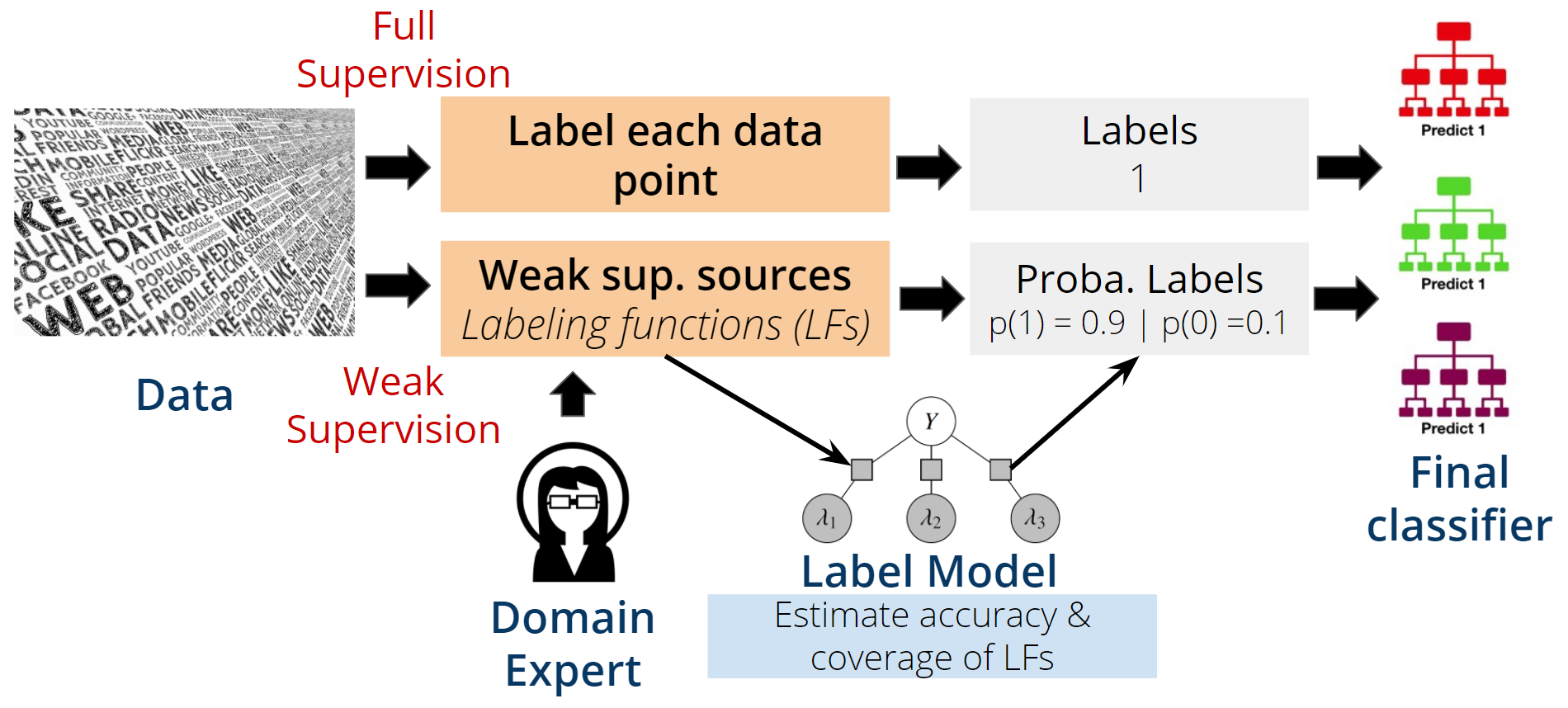}
    \caption{Data programming, or weak supervision compared to fully supervised ML. The orange boxes indicate the effort required by expert annotators. Instead of having to label extensive quantities of data by hand, the effort in data programming framework lies in obtaining labeling functions. In \texttt{KeyClass}, these labeling functions are our keyword-matching rules automatically extracted from reference data, to further reduce required human effort.}
    \label{fig:flowchart}
\end{figure}

The label model $\mathcal{L}_\theta$ assumes that each document is associated with an unobserved true class label $c_i \in \mathcal{C}$. Note that \citet{ratner2016data} defined the label model for multiclass but single-label classification. While this assumption holds for all our baseline datasets, it is not true for ICD-9 code assignment which involves multilabel classification. Hence, more generally, we assume that each document is associated with an unobserved set of labels, $\mathcal{S}^* \subseteq \mathcal{C}$, where $\mathcal{S}^*$ is a singleton set for single-label classification problems. 

To model the multilabel nature of the ICD code assignment, we further assume each document $d_i$ to be characterized by an \textit{unobserved} true probability distribution $p_i^*$ over the set of possible categories $\mathcal{C}$, an assumption which is consistent with literature on topic modelling \citep{blei2003latent}. Specifically, $p_i^*$ is a categorical distribution over all the categories $\mathcal{C}$, such that $p_i^*(c_j)$ is the probability that document $d_i$ belongs to class $c_j$. In a fully supervised multilabel classification, we would expect documents to be tagged with categories over which $p^*$ places a large probability mass.

The goal of the label model is then to label each document with $\hat p_i(c_j \mid \Lambda)$ given the votes of a set of $m$ labeling functions (LFs), $\Lambda = \{\lambda_i\}_{i=1 \dots m}$. For simplicity, the label model introduced by \citet{ratner2016data} assumes that all LFs are independent given the true class label, and that they vote with better than random accuracy where they do not abstain. However, the LFs do not need to have perfect accuracy and may conflict with one another.
We also assert that each LF only votes for a particular class by construction, i.e., we define each LF as $\lambda: \mathcal{D} \rightarrow c_i \in \mathcal{C}$. This allows us to use the same label model to estimate the accuracies and coverage of LFs using their agreements and disagreements via a factor graph, which is then used to infer a document's probabilistic label $\hat p_i(c_j \mid \Lambda)$, which is close to the true categorical distribution $p_i^*$ under settings enumerated in \citet{ratner2016data}, i.e.: $$ \forall d_i \in \mathcal{D}, \ \hat p_i(c_j \mid \Lambda) \approx p_i^*(c_j)$$ 

Let $\Lambda$ denote the $n \times m$ dimensional matrix of LF votes. In order to learn $\hat p(c_j \mid \Lambda)$, we first define a factor for LF accuracy as $\phi^{Acc}(\Lambda_{ij},c_i) \triangleq \mathbbm{1}\{\Lambda_{ij}=c_i\}$ as well as a factor of LF propensity as
$\phi^{Lab}(\Lambda_{ij},c_i) \triangleq \mathbbm{1}\{\Lambda_{ij}\neq 0\}$. Following \citet{ratner2016data}, we define the model of the joint distribution of $\Lambda$ and $C$ as:
{\small
\begin{align*}
    \label{eq:condind}
    & p_{\theta}(\Lambda,C) = \frac{1}{Z_\theta}\exp \left( \sum_{j=1}^m\sum_{i=1}^n \left( \theta_j \phi^{Acc}(\Lambda_{ij},c_i) + \theta_{j+m} \phi^{Lab}(\Lambda_{ij},c_i)\right) \right)
\end{align*}}
where $Z_\theta$ is a normalizing constant and $\theta$ are the canonical parameters for the LF accuracy and propensity. We use Snorkel~\citep{ratner2017snorkel} to learn $\theta$ by minimizing the negative log marginal likelihood given the observed $\Lambda$. Finally, we train a downstream classifier $\mathcal{M}_\phi$ with a noise aware loss function using the estimated probabilistic labels $\hat p_i( c| \Lambda)\}$.

\section{Methodology}
\begin{wrapfigure}[10]{r}{0.5\textwidth}
    \centering
    \includegraphics[width = 0.5\textwidth]{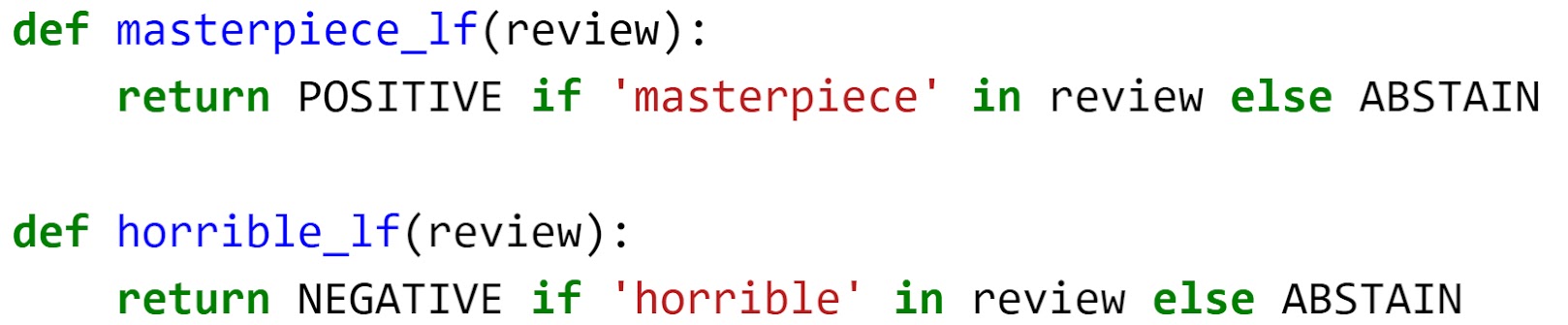}
    \caption{Example code of keyword labeling functions for the \texttt{IMDb} dataset. In practice, this is done automatically and implicitly through our pipeline.}
    \label{fig:lf_examples}
\end{wrapfigure}

\paragraph{Find Class Descriptions} 
Figure~\ref{fig:outline} presents an overview of our proposed method. Unlike traditional supervised learning where each document needs to be labeled \texttt{KeyClass} only relies on meaningful and succinct class descriptions. This also removes the requirement of expert heuristics as in prior weak supervision work.  
As a concrete example, consider the \texttt{IMDb} movie review sentiment classification problem, where the objective is to classify a movie review as being ``\textit{positive}'' or ``\textit{negative}''. In order to initiate the classification process, domain experts provide \texttt{KeyClass} with common sense descriptions of a positive (\texttt{"good amazing exciting positive"}) and negative review (\texttt{"terrible bad boring negative"}). Figure~\ref{fig:lf_examples} presents two example keyword labeling functions for the \texttt{IMDb} dataset.

In most cases, these descriptions can be automatically generated from Wikipedia articles or reference manuals and validated by domain experts, further reducing manual effort. For instance, for the ICD-9 code assignment problem, we can automatically acquire descriptions of all categories by mining the most frequently occurring words (minus stop words) from the combined text descriptions of all the codes (from CMS.gov\footnote{\url{www.cms.gov/Medicare/Coding/ICD9ProviderDiagnosticCodes/codes}}) that fall into each of our 19 categories. Table \ref{tab:example_mimic_descriptions} shows examples of the automatically curated descriptions of 2 high-level ICD-9 categories. Our approach overcomes a primary drawback of prior weak supervision methods, which rely on natural language rules manually crafted by domain experts.
\begin{table}[ht!]
    \centering
\resizebox{\textwidth}{!}{\begin{tabular}{c|c}
\Xhline{1pt}
\textbf{Category} &
  \textbf{Description}
   \\ \hline 
Respiratory system diseases &
  \begin{tabular}[c]{@{}l@{}} 
  due pneumonia acute chronic respiratory influenza pulmonary lung virus  asthma sinusitis bronchitis larynx \\ classified diseases obstruction elsewhere manifestations without identified pneumonitis ...
  \end{tabular}
  \\ \hline
Genitourinary system diseases &
  \begin{tabular}[c]{@{}l@{}} 
  specified chronic lesion female kidney acute glomerulonephritis disorders genital urinary cervix prostate breast ...
  \end{tabular} \\ \Xhline{1pt}

\end{tabular}}
\caption{Example descriptions for two ICD-9 categories. These descriptions were mined using the official descriptions of all the ICD-9 codes that fell into the ranges defined by each category \citep{world1988international}.}
\label{tab:example_mimic_descriptions}
\end{table}

\paragraph{Find Relevant Keywords} 
Once we have the class descriptions, \texttt{KeyClass} automatically discovers highly suggestive \textit{keywords} and \textit{phrases} for each class. Keywords have been shown to be excellent sources of weak supervision \citep{boecking2019pairwise, ratner2016data}. \texttt{KeyClass} first obtains frequent $n$-grams from the training corpus to serve as keywords or key-phrases for its automatically composed labeling functions. Let us denote the set of all keywords and key-phrases as $\mathcal{K}$. In our implementation, we used the \texttt{CountVectorizer} function from \texttt{scikit-learn} \citep{sklearn_api} to return \{1,2,3\}-grams having a document frequency strictly greater than $0.001$. We post-process the $n$-grams by removing common English stop-words from a corpus defined in the Natural Language Toolkit (NLTK) \citep{Loper02nltk}. 

In order to transform the keywords into labeling functions of the prescribed form, \texttt{KeyClass} leverage the general linguistic knowledge stored within pre-trained neural language models such as Bidirectional Encoder Representations from Transformers (BERT) \citep{devlin2018bert}, to map each keyword to the most semantically related category description. 
If we denote the neural language model encoder as $\mathbb{E} : \texttt{string} \rightarrow \mathbb{R}^d$, then at the end of this step \texttt{KeyClass} obtains two sets of equi-dimensional embeddings, $\{\mathbb{E}(k_i)\}_{i=1\dots |\mathcal{K}|}$ and $\{\mathbb{E}(e_i)\}_{i=1\dots c}$ corresponding to the keywords and class descriptions, respectively. Thus, to create a labeling function, \texttt{KeyClass} simply assigns a keyword to its closest category as measured by the cosine similarity between their embeddings. Thus, the keyword $k_i$ is assigned to its closest category $c_j$, when $j = \argmin_{\forall k \in c} d_\texttt{cosine}\left(\mathbb{E}(k_i), \mathbb{E}(e_k)\right)$. The corresponding labeling function can be then denoted as shown in Equation \ref{eqn:rule}. That is, if a document contains the keyword $k_i$ then, this labeling function votes for class $c_j$, otherwise it abstains from voting for any particular class.

For all our baseline experiments, we used \texttt{MPNET} \citep{song2020mpnet} to encode both the keywords/key-phrases and class descriptions into $768$-dimensional embeddings using the \texttt{paraphrase-mpnet-base-v2} implementation in the \texttt{sentence-BERT} \citep{reimers-2019-sentence-bert} Python package. For ICD-9 code assignment, we instead used \texttt{BlueBERT} \citep{peng2019transfer}, a BERT model trained specifically on clinical text databases, since it is better suited for clinical text classification problems such as ours. This modularity differentiates our method from previous methods such as \texttt{LOTClass} which are based on the specific neural language architectures (such as BERT) and hence inflexible to special problem domains.  

In order to ensure equal representation of all classes, \texttt{KeyClass} sub-samples the top-$k$ labeling functions per class, ordering them by cosine similarity. While theoretically data programming benefits from as many labeling functions as possible, the sampling is required due to computational and space constraints. For example, for the \texttt{AMAZON} and \textit{IMDb} datasets, we choose the top-$300$ labeling functions, whereas for \texttt{DBPEDIA} which has $14$ classes, we choose the top $15$ only. 

\paragraph{Probabilistically Label Data}
Next, \texttt{KeyClass} constructs the labeling function vote matrix $\Lambda$ and generates the probabilistic labels $\hat p(c_i \mid \Lambda)$ for all training documents using the label model $\mathcal{L}_\theta$ described earlier. Specifically, we use the open-source label model implementation of the \texttt{Snorkel} Python library released by \citet{ratner2016data}.

\paragraph{Train Downstream Text Classifier}
After obtaining a probabilistically labeled training dataset, \texttt{KeyClass} can train any downstream classifier using rich document feature representations provided by the neural language model $\mathbb{E}$. 
Instead of using all the automatically labeled documents, \texttt{KeyClass} initially trains the downstream classifier using top-$k$ documents with the most confident label estimates only. 

Finally, \texttt{KeyClass} self-trains the downstream model-encoder combination on the entire training dataset to refine the end model classifier. The primary idea of self-training is to iteratively use the model's current predictions $\mathbb{P}$ to generate a target distribution $\mathbb{Q}$ which can guide the model refinement using the following KL-divergence loss:
\begin{equation}
    \mathcal{L}_{ST} = \textbf{KL}(\mathbb{Q} || \mathbb{P}) = \sum_{i=1}^{n} \sum_{j=1}^{c} q_{ij} \log \frac{q_{ij}}{p_{ij}}
    \label{eqn:st_loss}
\end{equation}
where $p_{ij}$ is the predicted probability that the $i^{th}$ training sample belongs to the $j^{th}$ class. In order to compute the target distribution $\mathbb{Q}$, \texttt{KeyClass} applies soft-labeling which makes high confidence predictions more confident, and low confidence predictions less so, by squaring and normalizing the current predictive distribution $\mathbb{P}$ \citep{xie2016unsupervised}. More formally, 
\begin{equation}
    q_{ij} = \frac{p_{ij}^2 / f_j}{\sum_{j'}^K (p_{j'} / f_{j'})}, \ f_j=\sum_i^N p_{ij}
    \label{eqn:st_q}
\end{equation}

\section{Experiments}
\label{sec:experiments}

\subsection{Multilabel ICD-9 Code Category Classification}
In order to evaluate \texttt{KeyClass} on its ability to assign top-level diagnostic codes, we used free-text discharge summaries and corresponding ICD-9 codes recorded in the Medical Information Mart for Intensive Care (\texttt{MIMIC-III}) dataset \citep{johnson2016mimic}. \texttt{MIMIC-III} is a large publicly available single-center dataset comprising of de-identified clinical data of over $40,000$ patients admitted to the critical care units of the Beth Israel Deaconess Medical Center at Harvard Medical School between 2001 and 2012. For a faithful comparison of \texttt{KeyClass} with \texttt{FasTag}, we used the same $70:30$ train-test split and $19$ top-level ICD-9 categories used by \citet{venkataraman2020fastag}. 

Since this is a multiclass multilabel problem, we encode our target variable as 19-dimensional one-hot vectors, with a 1 corresponding to every diagnosis of a patient. While \texttt{KeyClass} does not require input text to be pre-processed, for consistency in comparing our model to \texttt{FasTag}, we follow \citet{venkataraman2020fastag}'s pre-processing by keeping only the most potentially useful for discrimination parts of text in each patient discharge note as ranked by the term frequency - inverse document frequency (TF-IDF) statistic. 

To compare our model against the supervised LSTM model in \texttt{FasTag}, we compute both aggregate \texttt{precision}, \texttt{recall}, and \texttt{F1} scores and category-specific \texttt{F1} scores as well as their confidences~\footnote{Specifically, we compute the \texttt{precision}, \texttt{recall} and \texttt{F1} scores for each instance and average across all test samples.}.

\subsection{General Weak Text Classification Performance of \texttt{KeyClass} Compared to Baselines}
\begin{table}[!ht]
\centering
\resizebox{0.75\textwidth}{!}{\begin{tabular}{ccccc}
\Xhline{1pt}
\textbf{Dataset} &
  \textbf{Classification Type} &
  \textbf{\# Classes} &
  \textbf{\# Train} &
  \textbf{\# Test}\\ \hline
\texttt{AGNews} & News topics & 4 & 120,000 & 7,600 \\ 
\texttt{DBPedia} & Wikipedia Categories & 14 & 560,000 & 70,000 \\ 
\texttt{IMDb} & Movie Reviews & 2 & 25,000 & 25,000 \\ 
\texttt{AMAZON} & Amazon Reviews & 2 & 3,600,000 & 400,000 \\
\texttt{MIMIC-III} & Clinical diagnostic categories & 19 & 39,541 & 13,181 \\ \Xhline{1pt}
\end{tabular}}
\caption{Dataset Statistics. All models are trained on the training set, but weakly supervised models do not have access to labels. Unlike other datasets, \texttt{MIMIC-III} is a multilabel multiclass classification problem where each clinical note must be assigned to all relevant categories. To best compare our results with prior work, we follow the same train and test splits as \citet{meng2020text} for the \texttt{AGNews}, \texttt{DBPedia}, \texttt{IMDb}, and \texttt{AMAZON} datasets. Similarly, for \texttt{MIMIC-III} we use the same train and test data as \citet{venkataraman2020fastag}. All datasets except \texttt{MIMIC-III} are balanced.}
\label{tab:data_summary}
\end{table}

We also compared \texttt{KeyClass} with previously proposed state-of-the art weakly supervised models (\texttt{Dataless} \citep{chang2008importance}, \texttt{WeSTClass} \citep{meng2018weakly} and \texttt{LOTClass} \citep{meng2020text}) and BERT-based fully-supervised \citep{devlin-etal-2019-bert} models on four real-world text classification problems. Since these previously proposed weakly supervised models were not tested on multilabel classification, we restricted our experiments to the following single-label multiclass problems: (1) movie review sentiment classification on \texttt{IMDb} \citep{maas-EtAl:2011:ACL-HLT2011} and \texttt{AMAZON} \citep{mcauley2013hidden}, (2) news topic classification on \texttt{AGNEWS} \citep{zhang2015character}, and (3) Wikipedia article classification on \texttt{DBPEDIA} \citep{lehmann2015dbpedia}. We used the same train-test splits as prior work (see Table \ref{tab:data_summary}) and report the accuracy accordingly. We also conducted ablation experiments to evaluate the impact of self-training and using data programming to probabilistically label the training data. 

We used a combination of a neural encoder and a $4$-layer MLP with LeakyReLU activations \citep{maas2013rectifier} as our downstream classifier. Each linear layer was followed by a dropout layer with $0.5$ dropout probability \citep{srivastava2014dropout}. To train the multilabel downstream classifier for ICD code assignment, we used the binary cross-entropy with logits loss. Cross-entropy loss was used to train classifiers for the remaining datasets. We trained each model with a batch size of $128$ for a maximum of 20 epochs, allowing for early stopping with a patience of $2$. We used Adam optimizer with learning rate of $0.001$\footnote{\textit{Justification of Modeling Decisions}: We consciously did not invest effort into optimizing hyper-parameters to not obfuscate the presentation of our core idea. But indeed, there is a good chance that the model could be further improved with manual or automated optimization.}. All models were built and trained using PyTorch $1.8.1$ \citep{NEURIPS2019_9015} using Python $3.8.1$. Experiments were carried out on a computing cluster, with a typical machine having $40$ Intel Xeon Silver $4210$ CPUs, $187$ GB of RAM, and $4$ NVIDIA RTX$2080$ GPUs.

\section{Results and Discussion}
\label{sec:resultsanddiscussion}
\subsection{\texttt{KeyClass} Assigns ICD-9 Codes Accurately.}
\begin{table*}[ht!]
\centering
\resizebox{\textwidth}{!}{
\begin{tabular}{c | c | c c c }
\Xhline{1pt}
\textbf{Supervision Type} & \textbf{Methods} & \textbf{Recall} & \textbf{Precision} & \textbf{F1} \\ \hline
Weakly sup. & \texttt{FasTag} \citep{venkataraman2020fastag} & $0.734\pm 0.00138$ & $0.436 \pm 0.00144$ & $0.525 \pm 0.00133$ \\
Weakly sup. & \texttt{KeyClass} (Ours) & $\textbf{0.896}\pm \textbf{.0009}$ & $\textbf{0.507}\pm \textbf{.0016}$ & $\textbf{0.6252} \pm \textbf{0.0014}$ \\ \hline
Fully sup. & \texttt{FasTag} \citep{venkataraman2020fastag} & $0.671 \pm 0.0019$ & $0.753 \pm 0.00171$ & $0.678 \pm 0.00141$ \\ 
\Xhline{1pt}
\end{tabular}}
\caption{\texttt{KeyClass} performs on par with the fully supervised baseline \texttt{FasTag} (Venkataramam et al.) on the \texttt{MIMIC-III} ICD-9 code assignment problem. We also report the performance of \texttt{FasTag} when trained using our probabilistic labels (weakly sup. \texttt{FasTag}). The superior performance of weakly supervised \texttt{KeyClass} over its \texttt{FasTag} counterpart is primarily due to better text modeling capabilities of \texttt{BlueBert} due to its relevant architecture and pre-training. The results are reported with 95\% bootstrap confidence intervals.}
\label{tab:MimicResults}
\end{table*}

\begin{table}[htb!]
\centering
\resizebox{0.65\columnwidth}{!}{\begin{tabular}{c| c c c}
\Xhline{1pt}
& \multicolumn{3}{c}{\textbf{Model performance}} \\ 
\textbf{Category Name} & \textbf{Prevalence} & \texttt{KeyClass} & \texttt{Fastag} \\ \hline
Infectious \& parasitic & 0.255 & 0.488 & \textbf{0.608}\\ 
Neoplasms & 0.157 & 0.031 & \textbf{0.656}\\ 
Endocrine, nutritional and metabolic & 0.626 & 0.855 & \textbf{0.862}\\ 
Blood \& blood-forming organs & 0.341 & \textbf{0.591} & 0.559\\ 
Mental disorders & 0.278 & \textbf{0.529} & 0.384\\ 
Nervous system & 0.232 & 0.329 & \textbf{0.499}\\
Sense organs & 0.068 & \textbf{0.004} & 0.002\\ 
Circulatory system & 0.760 & 0.922 & \textbf{0.936}\\ 
Respiratory system & 0.447 & 0.688 & \textbf{0.709}\\ 
Digestive system & 0.370 & 0.610 & \textbf{0.657}\\ 
Genitourinary system & 0.378 & 0.648 & \textbf{0.728}\\ 
Pregnancy \& childbirth complications & 0.003 & 0.000 & 0.000\\ 
Skin \& subcutaneous tissue & 0.107 & 0.004 & \textbf{0.090}\\ 
Musculoskeletal system \& connective tissue & 0.170 & \textbf{0.080} & 0.050\\ 
Congenital anomalies & 0.059 & 0.018 & \textbf{0.048}\\ 
Perinatal period conditions & 0.093 & 0.000 & \textbf{0.971}\\ 
Injury and poisoning & 0.347 & \textbf{0.608} & 0.601 \\ 
External causes of injury & 0.408 & \textbf{0.607} & --\\
Supplementary & 0.685 & \textbf{0.830} & -- \\ 
\Xhline{1pt}
\end{tabular}}
\caption{Performance comparison of \texttt{KeyClass} and \texttt{FasTag} disaggregated by ICD-9 categories, reveal that for some categories such as \textit{mental disorders} and \textit{injury and poisoning}, our model outperformed the fully supervised baseline. Both models had a hard time predicting low-support categories, i.e., categories with low prevalence in the data. Results for the last two categories were unavailable for \texttt{FasTag} \citep{venkataraman2020fastag}. }
\label{tab:mimic_f1}
\end{table}

Tables~\ref{tab:MimicResults} and~\ref{tab:mimic_f1} compare the performance of \texttt{FasTag} and \texttt{KeyClass} on the top-level ICD-9 code assignment problem. Remarkably, fully supervised \texttt{FasTag} achieves only $5$ points in F1 score over \texttt{KeyClass}, trained without any access to pointillistic labels or hand-coded natural language rules, with only minimal human intervention. 

Table \ref{tab:mimic_f1} reports F1 scores for each of the top level ICD-9 categories. Surprisingly, \texttt{KeyClass} outperformed \texttt{FasTag} on some categories such as \textit{mental disorders} and \textit{injury and poisoning}. We observed variance in the performance of \texttt{KeyClass} and \texttt{FasTag} across the categories. In fact, for some classes with high representation in the data, both models do well. On the contrary, the models report much lower F1 scores for less frequent classes.
We believe that further research is required to extensively analyze these lesser represented categories to ensure a high quality automated annotation system.

\subsection{\texttt{KeyClass} Outperforms Advanced Weakly Supervised Models.}
\begin{table*}[ht!]
    \centering
    \resizebox{\textwidth}{!}{
    \begin{tabular}{c | c | c | c | c | c} \Xhline{1pt}
    \textbf{Supervision Type} & \textbf{Methods} & \textbf{AG News} & \textbf{DBPedia} & \textbf{IMDb} & \textbf{Amazon} \\ \hline
    \multirow{4}{*}{Weakly sup.} & \texttt{Dataless} \citep{chang2008importance} &   0.696 & 0.634 & 0.505 & 0.501 \\
    & \texttt{WeSTClass} \citep{meng2018weakly} & 0.823 & 0.811 & 0.774 & 0.753 \\
    & \texttt{LOTClass} \citep{meng2020text} & 0.864 & 0.911 & 0.865 & 0.916 \\
    & \texttt{KeyClass} (Ours) & $\textbf{0.869} \pm \textbf{0.004}$ & $\textbf{0.940} \pm \textbf{0.001}$ & $\textbf{0.871} \pm \textbf{0.002}$ & $\textbf{0.928} \pm \textbf{0.000}$ \\ \hline
    Fully sup. & \texttt{BERT} \citep{devlin-etal-2019-bert} & 0.944 & 0.993 & 0.945 & 0.972 \\
    \Xhline{1pt}
    \end{tabular}}
    \caption{Classification Accuracy. \texttt{KeyClass} outperforms state-of-the-art weakly supervised methods on $4$ real-world text classification datasets. We report our model's accuracy with a $95\%$ bootstrap confidence intervals. Results for \texttt{Dataless}, \texttt{WeSTClass}, \texttt{LOTClass}, and \texttt{BERT} are reported from \citet{meng2020text}.}
    \label{tab:BaselineResults}
\end{table*}

Experiments on the \texttt{AGNEWS}, \texttt{DBPedia}, \texttt{IMDb} and \texttt{AMAZON} datasets reveal that \texttt{KeyClass} outperforms state-of-the-art weakly supervised models in terms of accuracy. Our model trained without access to any ground truth labels trails fully supervised \texttt{BERT} by less than $10$ percentage points (Table~\ref{tab:BaselineResults}). 

\subsection{Ablation Experiments: Self-training helps, Majority Vote is a strong baseline}
Table~\ref{tab:ablation} reports the results of our ablation experiments. Consistent with prior work, we observed varying degrees of improvement in downstream model performance from self-training. Self-training the downstream classifier improves its generalisation beyond the initial hypothesis learned from the top-$k$ most confidently labeled documents. We also found that taking the majority vote of labeling functions performs on par with data programming. In fact, on \texttt{Amazon} and \texttt{IMDb} majority vote outperforms data programming. This is most likely since \texttt{KeyClass} automatically creates a sufficiently large number labeling functions of approximately the same accuracy. On the other hand, in practice, data programming shines when there are few labeling functions with vastly different accuracies \citep{goswami2021weak}. 

\begin{table*}[ht!]
    \centering
    \resizebox{\textwidth}{!}{
    \begin{tabular}{c | c | c | c | c } \Xhline{1pt}
    \textbf{Methods} & \textbf{AG News} & \textbf{DBPedia} & \textbf{IMDb} & \textbf{Amazon} \\ \hline
    \texttt{LOTClass w/o. self train} \citep{meng2020text} & 0.822 & 0.860 & 0.802 & 0.853 \\
    \texttt{LOTClass} \citep{meng2020text} & 0.864 & 0.911 & 0.865 & 0.916 \\
    \texttt{\texttt{KeyClass} w/o. self train} & 0.$841 \pm 0.004$ & $0.823 \pm 0.002$ & $0.836 \pm 0.0019$ & $0.832 \pm 0.001$ \\
    \texttt{KeyClass} (Ours) & $\textbf{0.867} \pm \textbf{0.004}$ & $\textbf{0.951} \pm \textbf{0.001}$ & $\textbf{0.895} \pm \textbf{0.002}$ & $\textbf{0.941} \pm \textbf{0.00}$ \\
    \Xhline{1pt}
    Label Model (Data Programming) & \textbf{0.731} & \textbf{0.638} & 0.699 & 0.580 \\
    Label Model (Majority Vote) & 0.694 & 0.630 & \textbf{0.717} & \textbf{0.652}  \\
    \Xhline{1pt}
    \end{tabular}}
    \caption{Classification Accuracy for Ablation Experiments. Consistent with prior work, self-training improves downstream model performance. Data Programming performs on par compared to majority vote in probabilistically labeling the training data. Label Model accuracies are reported on the training set, whereas the rest of the results are reported on the test set. Results for \texttt{LOTClass} are reported from \citet{meng2020text}.}
    \label{tab:ablation}
\end{table*}

\subsection{Keywords are Excellent Sources of Weak Supervision.} Another finding of our study is that keywords and key-phrases are excellent sources of weak supervision. For more complex problems, it may be necessary for experts to manually re-assign some keywords to different categories. However, obtaining supervision at the keyword/key-phrase level is still much more efficient than labeling the entire corpus as any potentially required manual effort in our approach is upper-bounded by the size of the frequent terms vocabulary which is usually much smaller than size of the text corpus. 

\subsection{Limitations and Future Work.}
A limitation of \texttt{KeyClass} is that the automatic labeling function creation capability has not been tested on more complex problems. Moreover, our results on the ICD-9 code assignment problem show room for  improvement, especially in classifying low-resource categories. We believe that future work should focus on testing \texttt{KeyClass} on a wider range of complex real world text classification problems, and developing techniques to improve the performance of text classifiers on low support categories. 

One should take the labels predicted by \texttt{KeyClass} to be general areas where the clinical note is classified, not as definitive ground-truth--a human-in-the-loop situation would be best for practical applications. It would be interesting to see if our method can be useful for broadly checking the correctness of the already assigned ICD-9 labels in legacy results or reports, since their original hand labeling could yield errors. The resulting auditing tool could be beneficial for both the healthcare providers and insurers to improve accuracy of diagnostic coding, to reduce the risks of negative impact of such errors on quality of care and patient outcomes, and to mitigate the financial risks caused by coding errors in health insurance claims and reimbursement practices. 

\subsection{Conclusion}
Healthcare providers record detailed notes of clinical care delivered to each patient for clinical, research and billing purposes. Due to the unstructured nature of these narratives, providers employ dedicated staff to assign diagnostic codes to patients' diagnoses using the International Classification of Diseases (ICD) coding system. This manual process is, time-consuming, costly, and error-prone. 

The challenges are exacerbated by somewhat frequent revisions of the coding systems and their customization for use by particular healthcare organizations, so the currency and universality of the manual coding protocols are difficult to attain in practice, building up the costs and hassle. These challenges also limit practical utility of the existing, primary fully supervised machine learning approaches that rely on availability of substantial amounts of reference data to train reliable models for automated coding purposes.

To address this issue, we propose \texttt{KeyClass}, a general weak text classification model and a novel strategy to efficiently acquire interpretable weak supervision sources. \texttt{KeyClass} quickly and automatically creates highly interpretable heuristics based on keywords sourced from reference data, and enables end users to adapt its components to specific problems through support of domain specific language models. In contrast, previously proposed weakly supervised methods either rely on manually created heuristics, were uninterpretable due lack of transparency in the pseuo-labeling process, or were highly inflexible due their commitment to specific model architectures.

We successfully applied \texttt{KeyClass} to reliably assign ICD-9 codes over a large public dataset comprising of several thousand physician notes. We compared its performance with a state-of-the-art fully supervised model. We also found that \texttt{KeyClass} performs comparably, and for some code categories, even better than a supervised model trained using several thousand labeled clinical notes. Additional experiments on four standard NLP multiclass text classification problems confirm our proposed model's competitive position compared to previous methods. Although further research is necessary to comprehensively validate the proposed method across a wider range of complex data and use cases, \texttt{KeyClass}'s impressive performance on a challenging problem which plagues the healthcare industry, demonstrates its potential in helping broaden the adoption of beneficial machine learning technology in multiple application domains.





\acks{This work was partially supported by a fellowship from Carnegie Mellon University’s Center for Machine Learning and Health to M.G. The authors would also like to thank the anonymous reviewers and the program committee for their insightful feedback.}

\bibliography{references}

\end{document}